\pdfoutput=1

\documentclass[letterpaper, 10 pt, conference]{ieeeconf}  

\IEEEoverridecommandlockouts                              

\usepackage{graphicx} 
\graphicspath{{./img/}}
\usepackage{float}
\usepackage[labelformat=simple]{subcaption}

\usepackage{amsmath} 
\usepackage{amssymb}  
\usepackage{algpseudocode}

\title{Force-Guiding Particle Chains for Shape-Shifting Displays}

\author{
Matteo Lasagni and Kay R\"{o}mer
\thanks{Matteo Lasagni and Kay R\"{o}mer are with the Institute for \mbox{Technical} \mbox{Informatics,}
         Graz University of Technology, 8010 Graz, Austria
         {\tt\small \{lasagni,roemer\}@tugraz.at}}%
\thanks{Copyright is held by the author/owner(s).}
 }

\begin{document}

\maketitle
\thispagestyle{empty}
\pagestyle{empty}

\begin{abstract}
We present design and implementation of a chain of particles that can
be programmed to fold the chain into a given curve. The particles
guide an external force to fold, therefore the particles are simple
and amenable for miniaturization. A chain can consist of a large
number of such particles.  Using multiple of these chains, a
\emph{shape-shifting display} can be constructed that folds its
initially flat surface to approximate a given 3D shape that can be
touched and modified by users, for example, enabling architects to
interactively view, touch, and modify a 3D model of a building.
\end{abstract}

\section{Introduction}



The underlying goal of this work is the design and implementation of a ``shape
display'' -- a two-dimensional surface that can fold into the third
dimension where the shape to be displayed can be freely programmed
and dynamically changed. A user can not only view and touch the
displayed 3D surface, but by means of touch gestures recognized
through sensors built into the surface could interactively modify the
displayed shape. Such a shape display would enable a wide range of
interesting applications, for example, an architect could display and
interactively modify a newly designed building; scientists could
display complex 3D graphs and models to better understand them.

One approach to realize such a shape display is based on many tiny
modular robots (i.e., \emph{particles}) that can change their
arrangement such that the aggregate surface formed by all the robots
forms the displayed shape. However, approaches based on freely moving
autonomous particles have the disadvantage that each particle needs
complex actuators and latches to move into a desired configuration and
latch into a mechanically stable shape. Also, power supply and communication
among particles is challenging as they form dynamically
changing connection topologies. Therefore, researchers have
investigated approaches where neighboring particles are connected by
joints and can change their relative orientation by means of
actuators. The resulting fixed connection topology allows for wired
networking and power supply, but actuators and latches are still
needed in each particle.  Specifically, these actuators and latches
need to be strong enough to create large and mechanically stable
particle aggregations. This represents a hurdle towards
miniaturization of the particles, effectively limiting the
``resolution'' of the shape display to rather coarse structures.

Our contribution lies in removing the need for latches, only
requiring mechanically weak and simple actuators in each particle,
therefore enabling a better miniaturization of the particles. We call
our approach ``force-guiding particles'', as a single strong actuator
exerts a mechanical force on all particles in a chain and a weak and
simple actuator inside each particle guides this
force to fold the chain into a desired configuration. As the particles
in the chain are connected by joints, two wires running through the
chain provide power to and establish a communication network among all
particles. Therefore, miniaturization of particles becomes
feasible. By placing multiple such chains (all driven by a single
external actuator) next to each other we obtain a ``shape-shifting
curtain'' that can display 3D surfaces.

\begin{figure} 
  \begin{center}
      \includegraphics[width=0.4\textwidth]{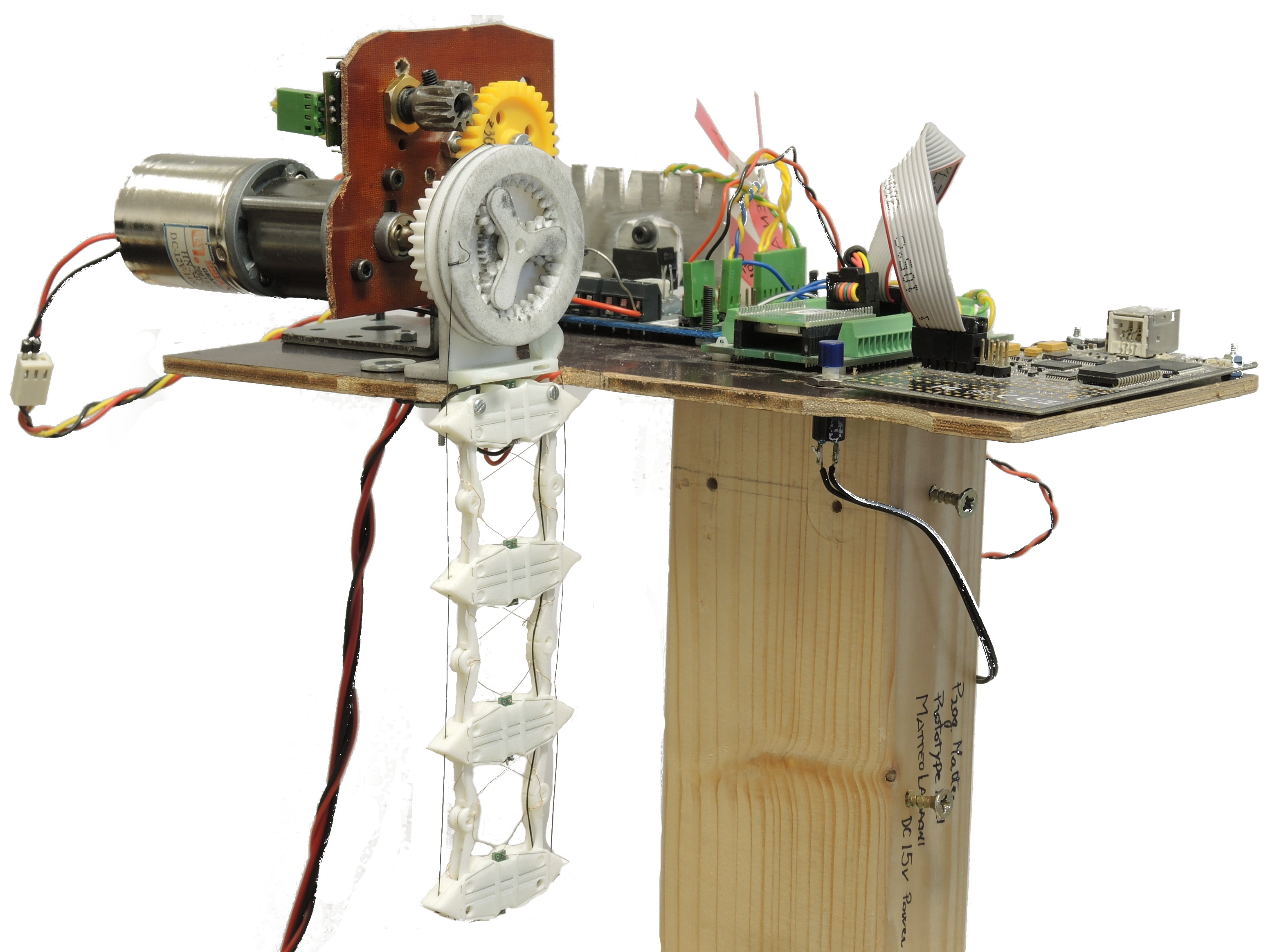}
      \caption{Our prototype of a ``force-guiding particle'' chain. 
      }
      \vspace{-6mm}
      \label{fig:prototype_overview}
  \end{center}
\end{figure}

The remainder of the paper describes the electro-mechanical design of
a force-guiding particle chain, as well as algorithms for control and
planning of chain folding to approximate a desired shape. We describe
a prototype (Fig.\ref{fig:prototype_overview}) that has been produced with a 3D printer and characterize
its performance. We also provide an analysis of the scaling properties
of force-guiding particle chains.

\section{Related Work}

``Shape-Shifting Material'' and ``Programmable
Matter'' are general terms for research on
materials whose physical properties and in particular shape can be
changed dynamically under program control or through direct
interaction \cite{goldstein:2005, weller:2011, ishii:2012}. The approach is
typically based on an extreme form of modular robotics, where small
robotic ``particles'' can change their relative position
by means of suitable actuators.

One can broadly identify two sub-classes of approaches. In the first
the robots are detachable and can ``climb'' each other
(e.g.,~\cite{jorgensen:2004, kirby:2007, murata:1998, zykov_2:2007,
gilpin:2008, romanishin:2013}).  While arbitary shapes can
be formed, the individual robots are typically complex as they need
complex actuators and latches as well as sophisticated power supply
and networking as the robots are not permanently connected. That leads
to costly robots which cannot be easily miniaturized. Even when
relocation relies on non-moving parts (e.g., magnets \cite{kirby:2007, kirby:2005}),
miniaturization might still be difficult as each particle needs to
excert strong forces on neighboring particles in order to form
mechanically stable shapes consisting in a large number of particles.

In the second sub-class ``robots'' are connected in fixed topologies
and the resulting substrate can be deformed by embedded actuators.
These substrates can be two-dimensional and fold along crease patterns
inspired by Origami (e.g., \cite{hawkes:2010, an:2011, onal:2011}),
or one-dimensional chains that fold in 2D (e.g., \cite{correl:2010})
or 3D (e.g., \cite{knain:2012, zykov:2007}). While this reduces the
complexity of the individual robots due to the fixed connection topology,
miniaturization of the individual robots is still difficult as the force
to fold and latch is still created by actuators inside each robot.

To overcome this intrinsic limitation, White et al. \cite{white:2010}
devise an external manipulator to fold a chain of passive latching
particles into 3D shapes. However, their external actuator is
rather complex (i.e., essentially a robotic arm) and a separate
actuator is needed for each chain, thus it does not scale well to
larger systems with multiple chains. Also, the unfolding of
the chain requires manual support. Hence, this approach is not
suitable for a ``shape display'' as we envision it where displayed
shapes can be dynamically changed.


\section{Requirements}

Below we explicate the main requirements on force-guiding particle chains.

%
\paragraph{Formation of arbitrary shapes}\label{req:arbitrary_shape_formation}
It should be possible to display a wide range of connected 3D surfaces. As our shape
display consists of multiple chains placed next to each other, each chain
should be able to approximate a 2D slice (i.e., curve) of the 3D surface.
\paragraph{Miniaturization of particles}\label{req:miniaturization}
To enable a high-resolution display, it should be possible to
miniaturize particles. This requires particles to have a
simple structure. In particular, it should be possible to miniaturize
the mechanical structures and actuators contained in a particle without
compromizing the mechanical stability of the particle chain.
\paragraph{Scalability}\label{req:scalability}
As particles are miniaturized, a growing number of particles are
required to display surface shapes of realistic size. Therefore, it should be possible to
include a large number of particles in each chain and to include many such chains into
a shape display, i.e., it should be possible to scale up the number of particles. Again,
mechanical structures and actuators play a key role here as they need to be strong enough
to deal with the growing extension and weight of the chain as the number of particles incrases.

 \section{Approach: Force-guiding Particles}



Our shape display consists of many particle chains placed
next to each other, thus forming a 2D surface. Each chain can fold within
a plane that is perpendicular to this surface. Together, all folded
chains approximate the 3D shape that is to be displayed.

Each chain resembles an articulated ladder-like skeleton
(Fig.~\ref{fig:mechanical_design}).  Here, a particle consists of the
square formed by two consecutive rungs and the two edges connecting
these rungs. By folding one of the edges, the square can be
reconfigured into a triangle, thus deflecting the chain. As each
particle in the chain can be folded left or right, the chain can
approximate any planar curve that does not self-intersect. Together, all
chains can thus approximate any connected 3D surface, thus meeting the
requirement of displaying arbitrary shapes. We outline a planning
algorithm to approximate a given shape later in the paper.

The regular geometry of the structure allows a strong external
actuator to exert a compressing force $F_w$ on the whole chain by
pulling two threads that run through the ladder structure on the left
and on the right as shown in Fig.~\ref{fig:mechanical_design}. By
exerting a small force $F_s$ on one of the edges it is slightly bent,
such that $F_w$ will compress the edge, thus transforming the
rectangular particle into a triangle. Edges to which no force $F_s$ is
applied remain locked in straight configuration even if $F_w$ is
applied. If $F_w$ is released, the chain returns into straight
configuration due to gravity. Thus, particles ``guide'' the external
force for folding.

The single external actuator (i.e., motor) is dimensioned such that it
can pull the threads of \emph{all} chains even if the number of
particles in each chain becomes large. While that may imply a
relatively big actuator, the dimensions of the particles (and thus the
resolution of the shape display) are not affected. In that sense, we
can scale up the number of particles in a chain without impact on the
size and complexity of the particles, thus meeting the scalability
requirement.

As we will explain in the following section, the force $F_s$
in the particle can be generated by means of a Shape Memory Alloy wire
that contracts when applying a current. By retaining the external
force $F_w$, the chain remains folded without the need for latches
inside each particle. Thus, the mechanics of a particle essentially
consists of hinges, joints, and SMA wires -- making it very simple
and enabling miniaturization, thus meeting the miniaturization requirement.

\section{Hardware and Software Design}
%
%


%
%
%
%

In this section we describe in more detail the design of force-guiding particle chains. We begin with
a description of the mechanics and electronics, followed by discovery and localization of particles,
and conclude with control and planning aspects.

\subsection{Mechanical Design}
\label{subsec:mechanical_design}

\begin{figure}
\vspace{0.5mm}
  \begin{subfigure}{0.22\textwidth}
    \centering
    \includegraphics[width=0.95\textwidth]{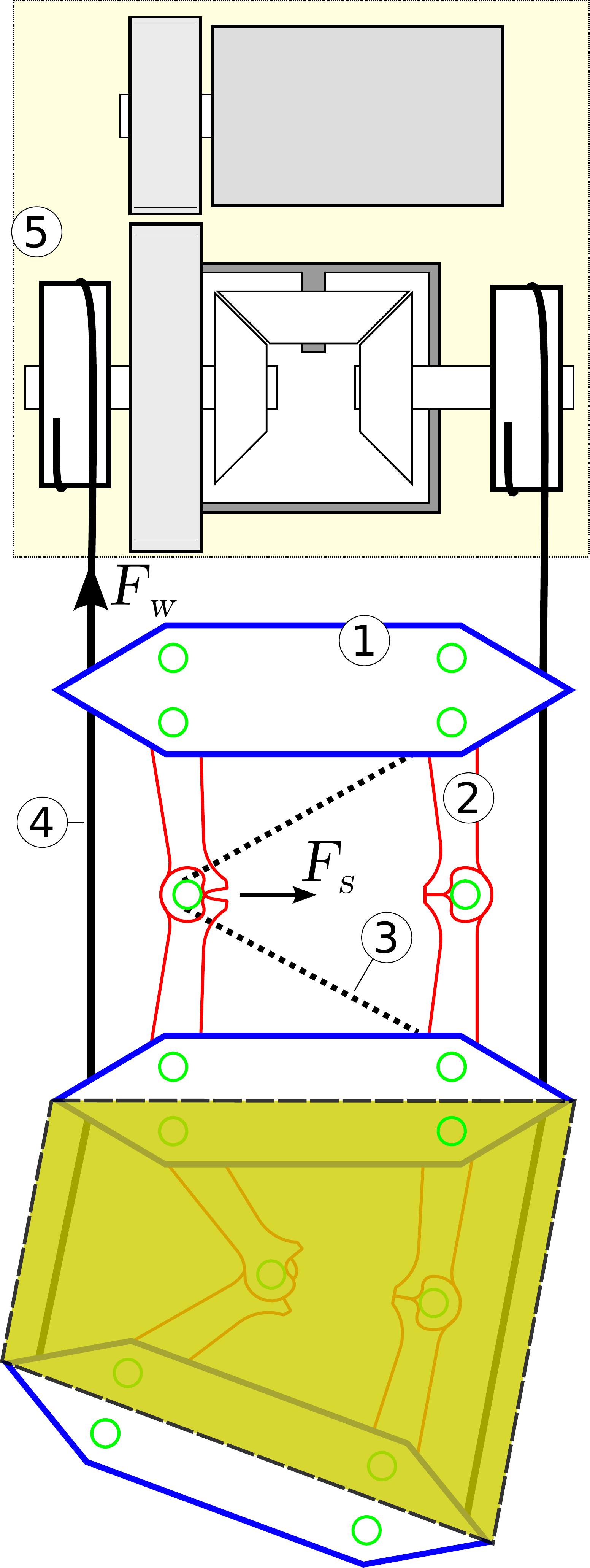} 
    \caption{Mechanical Design.}
    \label{fig:mechanical_design}
  \end{subfigure}%
  \hspace{5mm}
  \begin{subfigure}{0.22\textwidth}
      \begin{subfigure}{\textwidth}
        \centering
        \vspace{1mm}
     \includegraphics[width=.80\textwidth]{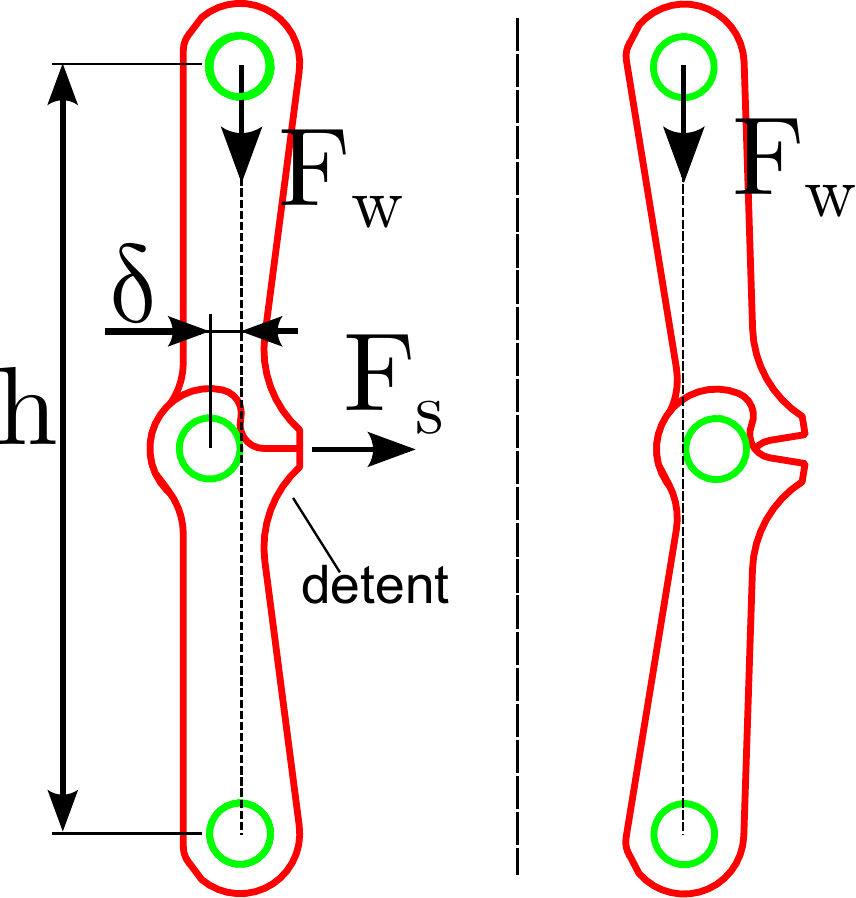}
      \caption{Detail of the flexible-edge.}
      \label{fig:monostable_link}
    \end{subfigure}
    \vspace{2mm}
    \begin{subfigure}{\textwidth}
     \includegraphics[width=.95\textwidth]{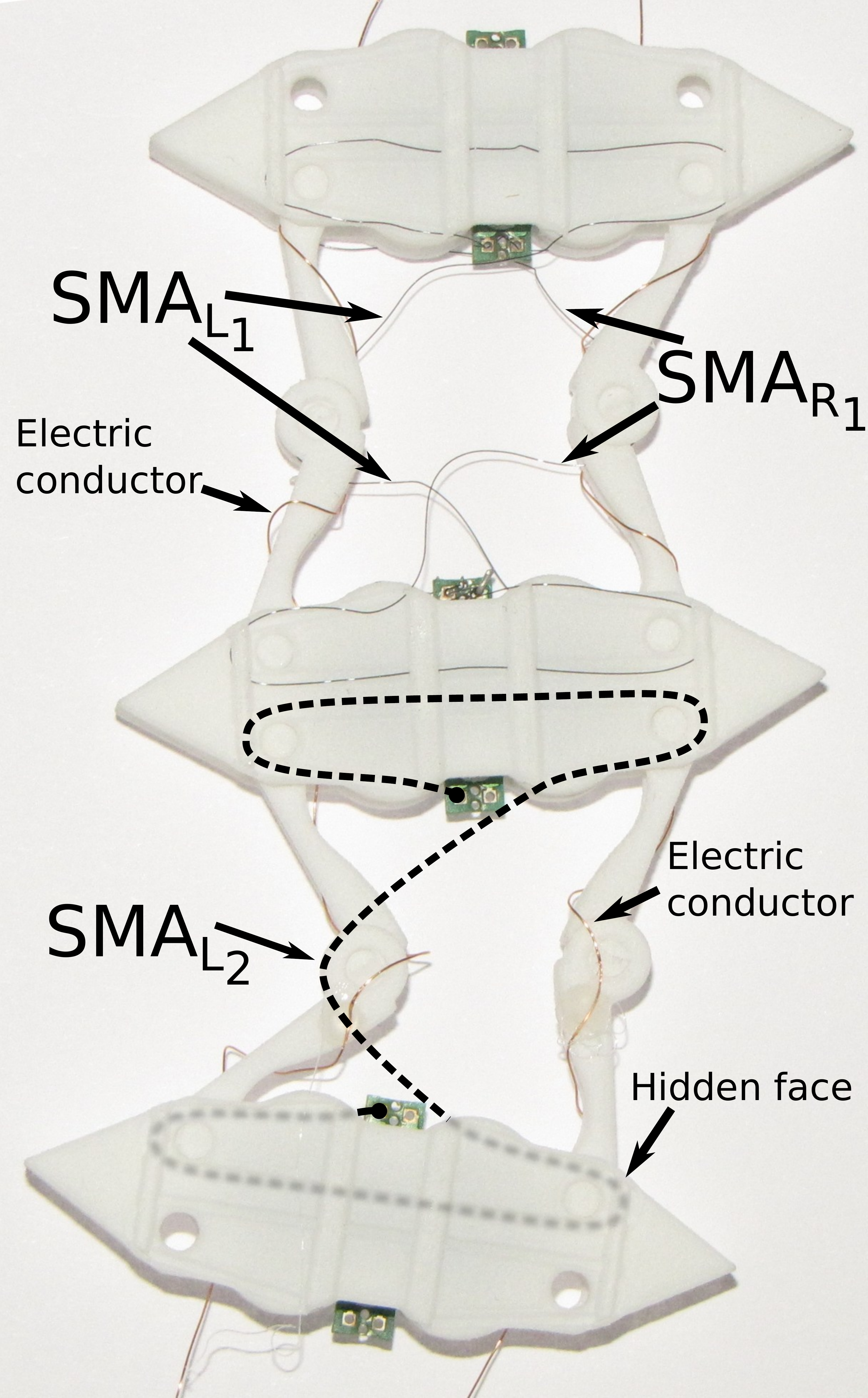}
      \caption{SMA path extension.}
      \label{fig:sma_path}
    \end{subfigure}
  \end{subfigure}
  \vspace{-2.5mm}
  \caption{Mechanical components of the system.}
  \vspace{-5mm}
  \label{fig:all_mechanics}
\end{figure}

Figure~\ref{fig:mechanical_design} shows the design of a particle
chain, where a deformable particle correspond to the highlighted
area. A particle chain consists of the following five main components
(see also the number labels in the figure).

\textbf{1. Rigid edge.}
    Each particle consists of two rigid horizontal edges that are shared with the particles above and below.
    The particular stretched-hexagon shape ensures a precise triangular mesh when the particles fold.
    Its dimensions are tailored to encase the electronics that control the particle.

\textbf{2. Flexible edge.}
    Hinged to successive rigid edges via snap-fit joints, each of the two flexible edges of a particle
    embodies a monostable mechanism to lock/unlock the folding.
    As shown in Fig.~\ref{fig:monostable_link}, the slight misalignment of the middle hinge w.r.t. the two hinges at the extremes,
    combined with a counteracting detent, lead to a ``locked'' straight configuration even when the external
    force $F_w$ is applied. When, instead, a small force $F_s$ pulls the middle hinge towards the
    centre of the particle, the link gets ``unlocked'', and $F_w$ deforms the particle into a triangular configuration.

\textbf{3. Active strap.}
    Made of Shape Memory Alloy (SMA), the contraction due to an electric current heating it up
    causes the lateral force $F_s$ to unlock the flexible edge as described above. When the current is
    stopped, the SMA cools down and returns to the original length.
    Since the SMA features a relative contraction of about $4\%$, a longer SMA wire running through
    the opposite rigid edges amplifies the contraction effect by a factor of four as shown in Fig.~\ref{fig:sma_path}.

\textbf{4. Actuation thread.}
    Next to the flexible edge, the actuation thread extends along the chain to propagate the force $F_w$
    the external actuator provides.
    The resulting effect on the chain is a compressing force on each side that, combined with the
    unlocking of a flexible edge, causes the deflection of the chain.
    By applying a constant tension, the chain retains the folded configuration without a need for latches.

\textbf{5. External actuator.}
    It provides the mechanical power to the system. An electric motor rolls up the two actuation
    threads, by means of spoolers connected via a differential gear.
    Because of the misalignment between flexible edges and threads, a particle deformation 
    makes one thread roll up, while the other one slightly unrolls. To mechanically compensate
    this effect, the differential gearbox decouples the two spoolers. In this way, a single motor
    can even drive multiple chains connected to the same shaft, provided an equivalent number
    of deforming particles among chains.

\subsection{Electronics}

\begin{figure} 
  \begin{center}
      \includegraphics[width=0.43\textwidth]{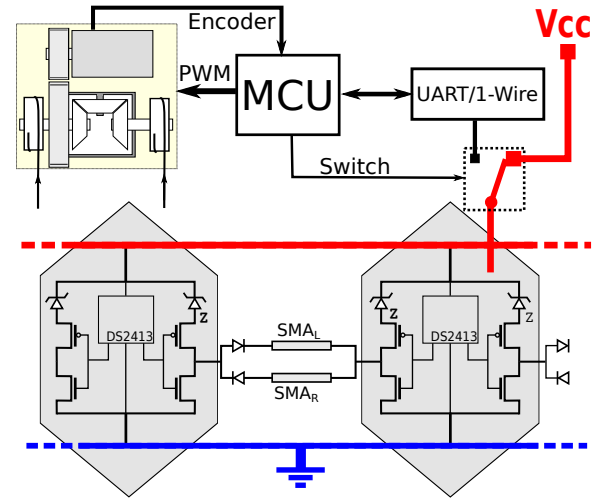}
    \caption{Electrical block diagram.}
    \vspace{-6.4mm}
     \label{fig:electrical_block_diagram}
  \end{center}
\end{figure}

Figure~\ref{fig:electrical_block_diagram} shows the electronic architecture
of the system, which consists of an MCU that controls the external actuator
as well as the particles via a 1-wire bus that also supplies power to the
particles.

\textbf{MCU.}
Core of the system is an 8-bit MCU (ATmega128RFA1) that controls
all the components: it remotely drives the local actuation of each particle
and synchronizes the external actuator to fold the chain.
As all the particles share the same bus to communicate with the
MCU and power the SMA straps, the MCU controls the bus line
switching it from communication mode (1-Wire\textsuperscript{\textregistered})
to SMA power supply mode (Vcc) and vice-versa.

\textbf{Actuation and feedback loop.}
The main actuator based on a 12\,V DC motor (HN-34PGD-2416T)
can rotate at variable speed (PWM controlled) in both directions
to fold and unfold the chain. The planetary gear reduction embedded in
the motor (gear ratio 410:1) prevents the backward rotation so
the motor can be switched off once the chain has been folded.
As no sensors are embedded in the particles, a quadrature
encoder (48\,cycle/rev.) connected through a gear (ratio 1:3)
to the main shaft, allows feedback control to stop the motor when a
desired configuration has been reached.

\textbf{1-Wire\textsuperscript{\textregistered} bus.}
Two conductors conveniently threaded through the chain
form the data line and ground reference for a 1-Wire\textsuperscript{\textregistered}
bus that connects particles and MCU.
1-Wire\textsuperscript{\textregistered} uses a master/slave protocol and
supports data and power on the same bus.
Power is supplied either when no communication is in progress or
when a logic '1' is transmitted. In these two cases, a pull-up resistor 
(in the master) pulls the voltage on the bus to 5\,V. While the logic
'1' passively results from the pull-up resistor, the dominant '0' is obtained
by short-cutting data line and ground. This limits the maximum current on
the bus to 5.4\,mA, as higher currents can introduce logic '0' in the
communication. Each 1-Wire\textsuperscript{\textregistered} slave exploits
the period when the bus is powered to harvest energy in a small capacitor.
As interface between the MCU and the bus, we adopted a
UART to 1-Wire\textsuperscript{\textregistered} converter (DS2480B)
that works as master.

\textbf{Particle/Node.}
Each particle hosts a dual-channel-addressable-switch (DS2413) encased in
the rigid edge of the particle which acts as a slave on the
1-Wire\textsuperscript{\textregistered} bus.
The MCU can control the output pins of these switches, with direct access to their
memory. As depicted in Fig.~\ref{fig:electrical_block_diagram},
each output pin drives a half-H bridge connected to a pair
of SMA straps: neighbouring nodes, connected to the same pair of SMA straps,
form a full-H bridge.
By controlling the polarity of the H-Bridge (i.e., one half-H bridge pulls up and
the other one pulls down) and introducing two diodes in mutual exclusion in series with the straps,
each strap ($SMA_L$ for left edge and $SMA_R$ for right edge)
can be activated independently.

\textbf{Bus commutator.}
As the maximum acceptable current on the 1-Wire\textsuperscript{\textregistered} bus
is 5.4\,mA, while the SMA straps nominally draw 180\,mA,
a commutator decouples data line and power supply (Vcc).
To obtain the same decoupling also within each particle,
a zener diode (5.6\,V) in series with the power circuit 
prevents the straps from being supplied when the bus
is in data mode. Instead, when the external commutator
switches the line to Vcc (15\textasciitilde20\,V), all the straps
previously enabled are powered.
The local capacitor on each DS2413 switch preserves the settings
during the voltage transition. Also, the device can withstand 28\,V.

\subsection{Node Discovery and Localization}
Each DS2413 is addressed by a factory lasered 64-Bit ID.  Since these
addresses are unknown until the device is connected to the bus, a
discovery algorithm is needed so that the MCU can find out the
addresses of the particles. The DS2480B bus master provides this
functionality.  However, the discovery algorithm does not indicate the
actual order of the particles along the bus. Since this is fundamental
for controlling the straps, we devise a localization algorithm to
infer the neighbor relationships among particles.


The algorithm exploits the fact that two neighboring particles
can control their half-H bridges such that a current flows through the SMA
wire connecting the two particles and the MCU can detect such a current
by measuring the power draw over the 1-wire bus.
%
%
%
The MCU hence picks one particle, configures its half H-bridge and
then sequentially selects all other particles (their addresses are known
after discovery) and configures their half-H bridge until power draw
varies, in which case two neighboring particles are found. The
procedure is repeated for the newly-found neighbor until no more
particles are left. If no neighbor can be found, a node is at the end
of the chain.

\subsection{Control}\label{subsec:control}

Folding the chain requires the MCU to synchronize a set of $N$ folding particles
$p_i$ and the main actuator. The process consists in four steps characterized
by time constants:

\begin{enumerate}
\item The 1-Wire master initializes the communication synchronizing the slaves while
discovering their addresses taking time $t_{setup}$;
\item The MCU sends the desired switch configuration to $k$ particles taking time $t_{config}$ for each;
\item The MCU switches the bus to power supply mode to activate the SMA taking a minimum time $t_{SMA}$ to fully contract the SMA;
\item The MCU activates the external actuator until the chain is folded taking time $t_{mot}$.
\end{enumerate}

If we assume that the folding operation involves $N$ particles, for which the configuration of
$k$ addressable switches needs to be set, the total time to perform the folding is:
%
\begin{equation}\label{eq:folding_time}
T(N, k) =  t_{setup} + k \cdot t_{config} + t_{SMA} + N \cdot t_{mot}
\end{equation}
%
%
As a result of the regular geometry of the chain, the number of revolutions the motor 
performs to deform each particle is constant. Then also $t_{mot}$ is constant for a constant speed.


\subsection{Planning}

Planning consists in computing for each particle in a chain whether it
should be folded left or right such that the folded chain approximates
a given curve S that is to be displayed.

\begin{figure} 
  \begin{center}
      \includegraphics[width=0.4\textwidth]{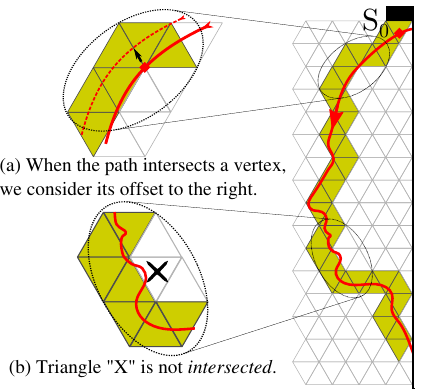}
      \caption{For each possible folding configuration of a chain, the particles lie
      on a regular triangular grid.}
  \vspace{-6mm}
      \label{fig:geometric_consideration}
  \end{center}
\end{figure}

We observe that for each possible folding configuration of a chain,
the folded triangular particles lie on a regular triangular grid as shown in 
Fig.~\ref{fig:geometric_consideration}, where the desired curve S resembles a
head. We further assume that the chain hangs on a wall where the top particle
is fixed to a wall bracket (black rectangle).

We choose the top most point $S_0$ of $S$, and from there start to
walk along $S$ to create
the sequence of triangles intersecting $S$. A triangle $t_i$ is said
to intersect $S$ if $S$ enters $t_i$ through edge $e_{i}$ and
eventually leaves $t_i$ through a different edge $l_{i} \neq e_{i}$.
For example, in zoomed Fig.~\ref{fig:geometric_consideration}(b), the
path intersects an edge of triangle ``X" and then leaves the
triangle through the same edge, therefore, the triangle marked ``X''
does not intersect.  In case the path $S$ intersects a vertex on the
grid, generating an undefined situation, we consider those triangles
that the path would intersect, if it had a slight offset to the right
(w.r.t the walking direction), as shown in
Fig.~\ref{fig:geometric_consideration}(a).
Each intersecting triangle $t_i$ corresponds to particle $p_i$ in the
chain. Likewise, each pair of edges $e_{i}$ and
$l_{i}$ corresponds to the opposite rigid edges of particle $p_i$.
Since we have chosen only adjacent intersecting triangles, the sequence of
corresponding particles is also connected.  To decide whether particle $p_i$
should fold left or right, we can walk again along the path
$S$ and consider the side where the corner connecting $e_{i}$ and $l_{i}$
lies: if it lies on the left of $S$, the particle folds the left edge; if it
lies on the right, the particles folds the right edge.

\section{Prototype}

%
%

We have built a prototype of a force-guiding particle chain 
consisting of three particles. All the components of the chain are
3D-printed. The resistance and the flexibility of the used material
PA22, along with the precision of the Selective-Laser-Syntering
technology allowed us to realize the parts of a particle with snap-fit
joints to simplify and speed up the assembly process. Specifically,
each rigid edge is composed of two symmetrical parts that can be
snapped into each other and encase the printed circuit board with the
electronics. Each half-link of a flexible edge is printed separately
and then snapped into the complementary part. The flexible edge is
then snapped into the two rigid edges. The SMA wire is threaded through
a thin tunnel (diameter 0.5\,mm) to extend the SMA path as shown in Fig.~\ref{fig:sma_path}.
We also investigated the possibility of printing pre-assembled hinges,
aiming to improve joint stability and minimize assembly overhead, but
the printing process requires a minimum distance between
surfaces which causes unacceptable backlash.

The differential gear and spoolers are also 3D printed. One specific
requirement for that component is that it should be thin enough to allow
placing multiple chains next to each other. For that reason, we designed
a planetary differential gear integrated with the spoolers.
The planet gears and the carrier are printed separately, then assembled
into the spoolers. Similarly, the gear connecting the motor and the
encoder is a custom 3D-printed component.

\section{Evaluation}

In this section we evaluate the design using the prototype described above.
We first perform an experimental validation of the prototype and report on
timing behavior and power consumption. We then analytically investigate the
fundamental scaling properties of the design.

\subsection{Experimental Validation}

A PC connected via USB/UART to the MCU lets the operator
control each particle individually and switches the bus between data
and Vcc. We firstly checked the correct functioning of the 1-wire
network, showing that the data line can be switched to Vcc for powering
the SMA wires without issues for communication. We found that the 1-wire
master looses synchronization with the slaves after this operation,
requiring a reset to restore the communication.  This introduces a
delay after each operation equivalent to $t_{setup}$
(Sec.~\ref{subsec:control}). We measure that $t_{setup}$ takes 205\,ms
with 6 nodes for the initialization of the bus; later resets can be performed
within 78\,ms, regardless of the number of nodes on the bus.  The configuration
of a node takes $t_{config}$=22\,ms.

We installed a chain of three particles hanging from a custom support
with the main actuator winding up the thread from the top.
We empirically measured the activation time of the SMA to fully disengage
the link (using video), and counted the actual revolutions of the spool
to completely fold one particle. The SMA requires $t_{SMA} \ge$\,421\,ms to disengage the
link when the system is supplied with 20\,V and the SMA
draws 280\,mA (nominal value 180\,mA). However, for safety reasons we
increased this value to 500\,ms.

Considering that to completely fold the chain the thread winds up by
nominally 99.1\,mm, the spooler (40\,mm nominal diameter) requires
0.789 revolutions, which corresponds to approximately 113 cycles on the
encoder (48 cycles/rev. connected to the shaft through a 1:3 gear).
We reduce the motor speed using pulse width modulation.
The time needed to actuate a single particle is then
$t_{mot}$=1050\,ms.

Inserting these values into Eq.\ref{eq:folding_time} and considering
a single particle folding (i.e., two addressable switches have to be
to set), the time needed for this operation is $T(1, 2) = 78
\mathrm{ms} + 2 \cdot 22 \mathrm{ms} + 500 \mathrm{ms} + 1 \cdot 1050
\mathrm{ms} = 1672 \mathrm{ms}$, which matches the experimental result
we obtain.  Actuating the system with the motor at the nominal speed
of 14\,RPM, $t_{mot}$ reduces to 56\,ms, for which $T(1, 2)$
becomes 678\,ms. The unfolding operation only requires 168\,ms with
three particles.


%


A complete actuation test, where all the particles are folded starting from the 
bottom and progressively proceeding to the top, demonstrated the correctness
of the settings previously identified. Even though each deformation presents
an error of $\pm$2 cycles on the encoder (mainly due to fabrication tolerance,
but also due to the low accuracy of the measurements), the errors self-compensate
when all the particles are folded.

%

The unfolding of the chain subject to gravity could not completely
restore the initial configuration, although an additional rubber bands
assist links to return to straight configuration. This
happens because of the moderate weight of the components and
especially due to the friction at the hinges. An additional weight of
16\,g added to the tail of the chain improves the result, yet
without completely unfolding the chain. Improvements in the manufacturing
process to reduce the friction will address this issue.

\subsection{Scalability}

\begin{figure}
  \begin{center}
        \includegraphics[width=0.40\textwidth]{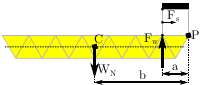}
      \caption{Worst case configuration of a folded chain.}
       \vspace{-6mm}
      \label{fig:cantilever}
  \end{center}
\end{figure}

In this section we investigate the scaling limit of our approach,
i.e., how many particles we can support in a single chain. For that we
consider the worst-case situation depicted in
Fig.~\ref{fig:cantilever}, where $N$ particles of a chain are
already folded in a straight horizontal configuration (cantilever),
while the $(N+1)$-th particle (top-right in the figure) is about to fold
by applying $F_s$ to the left flexible edge.  We estimate the force
$F_w$ resulting on the pulling thread and compute the maximum number
$N$ of particles that can be supported such that the limited force
$F_s$ that can be exerted by the SMA is still sufficient to bend the
flexible edge given $F_w$. Due to the leverage effect $F_w$ is
maximized in the shown configuration, therefore it constitutes the
worst case. For this we first have to compute the force $F_s$ that is required to
bend a flexible edge when a pulling force $F_w$ is applied to the particle:
\vspace{-2mm}
\begin{equation}\label{eq:mech_advantage}
 F_s = 2\ \tfrac{\delta}{h}\ F_w
\end{equation}
where $\delta$ is the offset between the three hinges of the
monostable link and $h$ the distance between the two hinges at the
extremes as shown in Fig.~\ref{fig:monostable_link}.

Inserting the actual dimensions of our prototype into
Eq.~\ref{eq:mech_advantage}, we can estimate the mechanical advantage
$F_w/F_s$. With $\delta$=1.28\,mm and $h$=32.11\,mm, the
mechanical advantage is 12.5. A smaller $\delta$ would increase this
value, but would also reduce the stability of the flexible edge.

Now we can establish a relationship between the number $N$ of
particles and the resulting force $F_w$. For this we observe that the
configuration shown in Fig.~\ref{fig:cantilever} can be considered a
lever of class 3, where $P$ is the fulcrum, $F_w$ the effort, and
$W_N$ the resistance.  We define $W_N$ as the total weight of the
cantilever applied to its centre of mass, $W_N = N W_p$, where $W_p$
is the weight of one particle (about 4.5\,g in our prototype). The two
distances $a$ and $b$ correspond to the edge length $l$ of a particle
and $\tfrac{1}{4}N l$, respectively. Applying the law of the lever we obtain:
\vspace{-2mm}
\begin{equation}\label{eq:strength_cantilever2}
  F_w \cdot l = \tfrac{1}{4} N l \cdot (N W_p)
\end{equation}
We can now substitute $F_w$ with $12.5 \cdot F_s$ according to the
mechanical advantage computed above based on Eq.~\ref{eq:mech_advantage}
and solve for $N$:
\vspace{-2mm}
\begin{equation}\label{eq:number_of_particle}
  N  = 2 \sqrt{\frac{F_w}{W_p}}=2 \sqrt{\frac{12.5\cdot F_s}{W_p}}
\end{equation}
The maximum $F_s$ an SMA strap can exert, depends on its contraction
force: with a diameter of 100\,$\mu$m the nominal contraction force
$F_{SMA}$ is 14.7\,N. As the arrangement of the strap around the hinge
of the flexible edge (Fig.~\ref{fig:all_mechanics}) doubles the effect
of its contraction force, while the angle the strap forms with the
vector $\overline{F_s}$ reduces its effect by $cos(30^\circ)$, we can
approximate $F_s = 2 F_{SMA} \cdot cos(30^\circ)$ which corresponds to
2.59\,kg. Substituting this value and $W_p=4.5$\,g into
Eq.~\ref{eq:number_of_particle}, we obtain $N=84$ particles in the worst case.
We observe that $N$ is independent of the size $l$ of the particles
and grows with the inverse of the square root of the weight $W_p$ for a
constant $F_s$. This means that if we can miniaturize particles and reduce
their weight, we can increase the number of particles in the chain in the worst case.
Hence, the miniaturization is only limited by the mechanical stability of the
thread and the material from which the particles are made.

\section{Conclusion And Future Work}

We have presented a force-guiding particle chain that can fold into
arbitrary flat curves under program control as a building block to
construct a shape-shifting display that can fold its surface into a 3D
shape that can be viewed, touched, and modified by users. The key
feature of force-guiding particle chains is that the mechanical force
to fold the chain is generated by an actuator external to the chain,
such that the particles of the chain are simple and require only a
mechanically weak actuator to guide the external force. Thereby,
particles are amenable to miniaturization and scale up to a large
number of particles per chain. We presented the mechanical and
electrical design, as well as algorithms for control and planning. We
demonstrate and validate a working prototype.

Future work includes, among others, improvement of the mechanical
design to reduce friction and the number of parts; inclusion of sensors into
particles for recognizing touch gestures and folding state of particles;
as well as advanced planning algorithm to minimize $F_w$.

\bibliographystyle{IEEEtran}
\bibliography{ref}

\end{document}